\documentclass[10pt,twocolumn,letterpaper]{article}

\usepackage[pagenumbers]{cvpr} 

\definecolor{cvprblue}{rgb}{0.21,0.49,0.74}

\def\paperID{34722} 
\def\confName{CVPR}
\def\confYear{2026}

\usepackage{booktabs}   
\usepackage{multirow}   
\usepackage{graphicx}   
\usepackage{lipsum} 
\usepackage{booktabs}
\usepackage{graphicx}

\usepackage{amssymb}
\usepackage{enumitem}
\usepackage[table]{xcolor}
\usepackage{array}   
\usepackage{caption}
\usepackage[utf8]{inputenc}
\usepackage{tcolorbox}
\tcbuselibrary{skins,breakable} 
\newcolumntype{C}[1]{>{\centering\arraybackslash}m{#1}}
\usepackage[pagebackref,breaklinks,colorlinks,allcolors=cvprblue]{hyperref}
\usepackage{bookmark}

\title{ EgoMind: Activating Spatial Cognition through Linguistic Reasoning in MLLMs}

\author{
    Zhenghao Chen$^{1,2}$ \quad Huiqun Wang$^{1,2}$ \quad Di Huang$^{1,2}$\thanks{Corresponding author.} \\
    $^1$State Key Laboratory of Complex and Critical Software Environment, Beihang University \\
    $^2$School of Computer Science and Engineering, Beihang University \\
    {\tt\small \{zhenghao.chen, hqwangscse, dhuang\}@buaa.edu.cn}
}

\begin{document}
\maketitle
\begin{abstract}

Multimodal large language models (MLLMs) are increasingly being applied to spatial cognition tasks, where they are expected to understand and interact with complex environments. Most existing works improve spatial reasoning by introducing 3D priors or geometric supervision, which enhances performance but incurs substantial data preparation and alignment costs. In contrast, purely 2D approaches often struggle with multi-frame spatial reasoning due to their limited ability to capture cross-frame spatial relationships. To address these limitations, we propose \textbf{EgoMind}, a Chain-of-Thought framework that enables geometry-free spatial reasoning through Role-Play Caption, which jointly constructs a coherent linguistic scene graph across frames, and Progressive Spatial Analysis, which progressively reasons toward task-specific questions. With only 5K auto-generated SFT samples and 20K RL samples, EgoMind achieves competitive results on VSI-Bench, SPAR-Bench, SITE-Bench, and SPBench, demonstrating its effectiveness in strengthening the spatial reasoning capabilities of MLLMs and highlighting the potential of linguistic reasoning for spatial cognition. Code and data are released at \url{https://github.com/Hyggge/EgoMind}.
\end{abstract}

\section{Introduction}
\label{sec:intro}

With the rapid advancement of multimodal large language models (MLLMs), these models have been increasingly adopted in embodied intelligence, virtual reality (VR), and augmented reality (AR). Consequently, enhancing their spatial reasoning capabilities has become critical for enabling intelligent perception, reasoning, and interaction within complex environments.

To this end, most concurrent approaches integrate explicit 3D inputs into MLLMs pretrained on vision–language modalities. Researchers have explored diverse 3D information sources, including 3D point clouds~\cite{Chat3D, LL3DA, LEO, ChatScene, SIG3D, Grounded-3D-LLM, Spatial3DLLM}, bird’s-eye-view (BEV) representations~\cite{GPT4Scene, Struct2D}, depth maps~\cite{SpatialPIN, MM-Spatial, GSReasoner}, camera parameters~\cite{3D-LLM, Scene-LLM, LLaVA-3D}, egocentric trajectories~\cite{See_Trek}, and geometric features~\cite{3DRS, 3DThinker, VLM-3R, Spatial-MLLM} distilled from pretrained 3D backbones~\cite{VGGT, CUT3R}. Facilitated by these geometry-based priors and alignment strategies, MLLMs acquire a global, metrically consistent understanding of scenes, achieving stronger spatial comprehension than vanilla counterparts.

\begin{figure*}[t]
  \centering
  \includegraphics[width=1\linewidth]{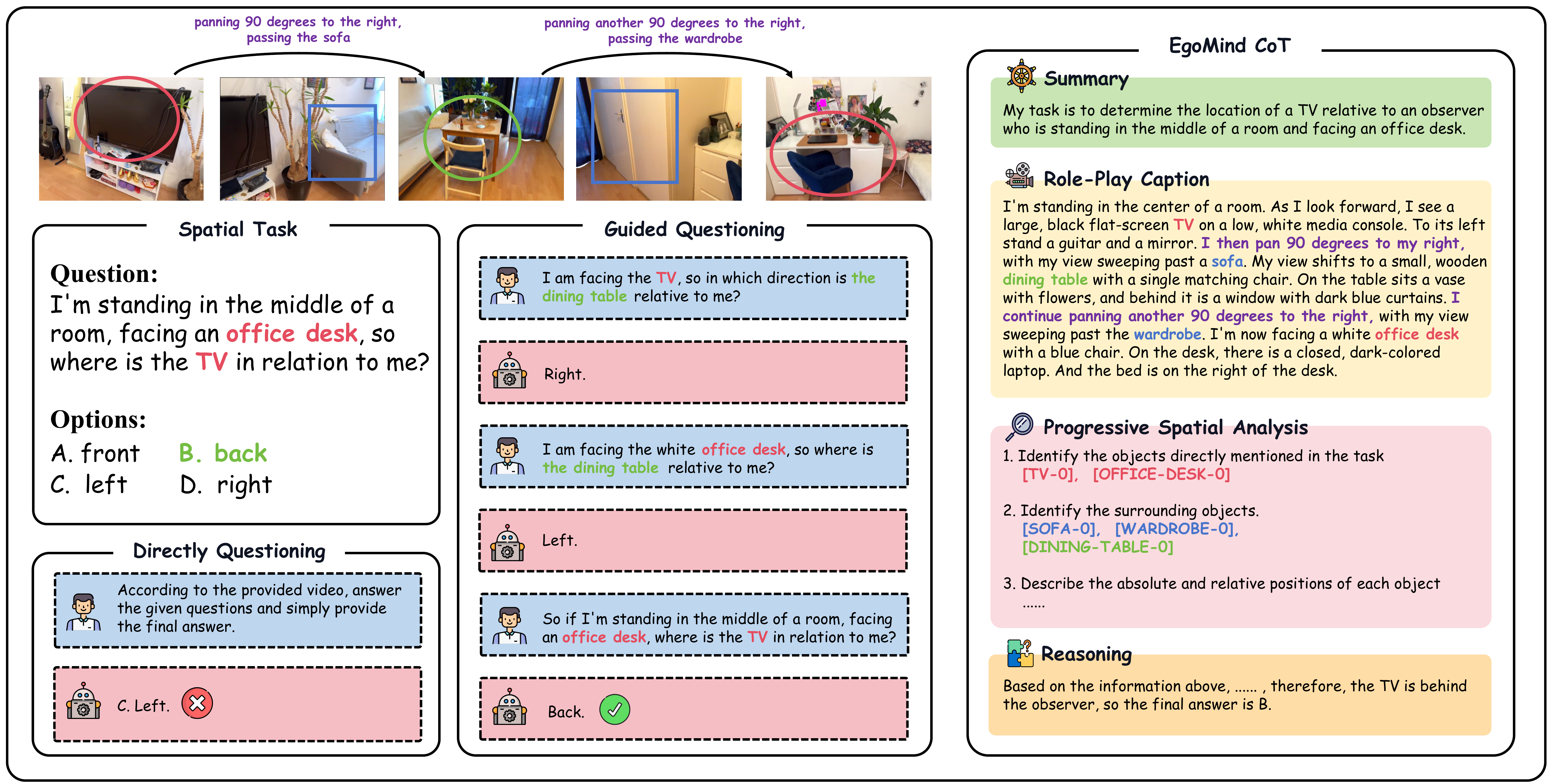}
   \caption{Illustration of the differences among spatial reasoning approaches. Direct questioning often fails because of missing cross-frame correlations and limited awareness of implicit objects needed for spatial bridging. Guided questioning helps the model gradually establish these associations. In contrast, EgoMind CoT explicitly models viewpoint transitions and implicit spatial bridges, builds a coherent global scene representation, and reliably produces the correct answer.}
   \label{fig:problem}
   \vspace{-0.5cm}
\end{figure*}

Despite these advancements, such approaches rely heavily on additional modalities or geometric supervision. Integrating 3D priors into MLLMs typically requires paired multimodal data~\cite{Chat3D, LL3DA, LEO, ChatScene, SIG3D} or geometry-guided projection of 2D features into 3D space~\cite{3D-LLM, Scene-LLM, LLaVA-3D} during pretraining. Moreover, additional embedding alignment mechanisms~\cite{3DRS, 3DThinker, VLM-3R, Spatial-MLLM} are needed to fuse heterogeneous 2D and 3D representations. These requirements introduce substantial data preparation and alignment burdens, leading to high training costs and limited generalization. A few recent studies~\cite{SpaceR, SpatialVLM, Video3DLLM} have sought to enhance spatial reasoning using purely image--language inputs. Nevertheless, despite the strong performance of recent models on single-frame scene understanding~\cite{Qwen25VL, Ovis2.5, Intern-S1, InternVL3.5, MINICPM-V4.5, Kimi-VL}, these approaches still struggle to internalize spatial understanding in complex multi-frame scenarios. Without explicit 3D priors, MLLMs must infer the underlying spatial structure solely from 2D frames. However, as illustrated in Fig.~\ref{fig:problem}, they often fail to establish the cross-frame spatial associations required for accurate reasoning: direct inference frequently leads to incorrect answers due to disrupted viewpoint continuity and limited awareness of implicit objects that act as spatial bridges. Although handcrafted progressive prompting can sometimes steer the model toward a correct reasoning chain, such a process is neither scalable nor reliable.

To further examine this gap, we identify two underlying challenges. First, many existing methods process multi-view inputs on a frame-by-frame basis, without explicitly modeling the continuous spatio-temporal transformations and geometric relationships across viewpoints, resulting in fragmented cross-frame spatial understanding. Second, MLLMs tend to focus exclusively on objects explicitly mentioned in the query, while overlooking implicit yet crucial contextual elements needed to bridge observations across frames, resulting in incomplete or erroneous reasoning chains. Together, these issues hinder pure vision--language MLLMs from constructing coherent spatial representations and performing robust multi-frame reasoning.

To address these challenges, we enhance MLLMs' spatial reasoning through carefully structured linguistic signals, enabling them to bridge cross-frame viewpoint discontinuities and reason more effectively about implicit object relations. Specifically, the Role-Play Caption (RPC) component simulates an agent navigating an environment from a first-person perspective, generating coherent descriptions of frame-wise observations and viewpoint transitions to build a consistent global understanding of the scene. In parallel, the Progressive Spatial Analysis (PSA) component first localizes objects explicitly mentioned in the query, then expands its attention to surrounding entities, and finally reasons about their spatial relationships in an integrated manner. By combining these two components, we propose a novel chain-of-thought (CoT) framework, termed EgoMind, that jointly models inter-frame dependencies and implicit object relations, thereby substantially improving spatial understanding without relying on geometric inputs or explicit 3D priors.

Benefiting from the flexibility and abstraction capabilities of linguistic reasoning, EgoMind remains highly cost-efficient to train while delivering strong spatial reasoning performance. Using only 5K automatically generated samples for supervised fine-tuning, without handcrafted annotations, and 20K samples for reinforcement learning, EgoMind achieves competitive results on VSI-Bench~\cite{VSI-Bench}, SITE-Bench~\cite{SITE-Bench}, SPBench~\cite{Spatial-Ladder}, and SPAR-Bench~\cite{SPAR-Bench} under purely image--language supervision, demonstrating both the efficiency and effectiveness of the proposed framework.

Our contributions are as follows:

\begin{itemize}
    \item We propose EgoMind, a novel framework with a specially designed CoT paradigm that integrates Role-Play Caption and Progressive Spatial Analysis to induce spatial understanding through linguistic reasoning.
    
    \item We develop a fully automated data generation pipeline based on the EgoMind CoT formulation, requiring no human annotation and enabling cost-efficient training with reduced data overhead.
    
    \item Extensive experiments on the benchmarks show that EgoMind achieves competitive performance among open-source MLLMs, validating the proposed framework.
\end{itemize}

\section{Related Work}
\label{sec:relative_works}

\subsection{Multimodal Understanding and Reasoning}

Recently, MLLMs have advanced rapidly, exhibiting increasingly strong capabilities in multimodal understanding and reasoning. Early efforts focused on architectural design. For example, BLIP-2~\cite{BLIP2} introduced the Q-Former, while Flamingo~\cite{Flamingo} employed cross-attention to build a unified vision--language embedding space. The LLaVA series~\cite{LLaVA, LLaVA-1.5, liu2024llavanext} further established an LLM-centric paradigm in which visual inputs are projected into the language embedding space through a simple MLP. Owing to its simplicity and effectiveness, this paradigm has been widely adopted in subsequent MLLM research.

Building on this foundation, training strategies such as multi-stage alignment~\cite{li2024llava,  Qwen25VL, InternVL3.5, Ovis2.5} and instruction tuning~\cite{zhu2023minigpt4, chen2023sharegpt4v}, together with large-scale visual instruction datasets~\cite{li2024llava, tong2024cambrian}, have been proposed to develop stronger open-source multimodal models~\cite{li2024llava, Qwen25VL, Ovis2.5, Intern-S1, InternVL3.5, MINICPM-V4.5, Kimi-VL}. To further enhance multimodal reasoning, LLaVA-CoT~\cite{LLaVA-CoT} structures reasoning into four stages for step-by-step inference, while Mulberry~\cite{Mulberry} leverages a collective Monte Carlo tree search to learn from explicit reasoning trees. In addition, reinforcement learning methods inspired by DeepSeek-R1~\cite{Deepseek-R1} have been introduced to strengthen general reasoning~\cite{Visual-RFT, R1-Onevision, Vision-R1, Video-R1, Skywork-R1V, R1-VL, ORZ}, further pushing the reasoning capabilities of MLLMs.

Despite these advancements, current MLLMs still struggle with spatial understanding and reasoning when relying purely on image--language inputs. In particular, capturing spatial relationships and maintaining a coherent global perception across multiple views remain open challenges.

\subsection{Spatial Understanding and Reasoning}

Driven by the growing application of MLLMs in spatial cognition tasks, recent studies have introduced various strategies to enhance spatial understanding capabilities.

3D prior–based approaches focus on integrating explicit 3D information into MLLMs to improve spatial reasoning and scene comprehension. LL3DA~\cite{LL3DA} and LEO~\cite{LEO} employ additional 3D branches to incorporate point clouds for enhanced scene-level understanding. Grounded 3D-LLM~\cite{Grounded-3D-LLM} designs a cross-modal interaction module to improve fine-grained object reasoning in 3D space, while Chat3D~\cite{Chat3D} and ChatScene~\cite{ChatScene} utilize 3D detectors or segmentors to extract explicit object features from 3D modalities. Beyond point clouds, 3D-LLM~\cite{3D-LLM}, Scene-LLM~\cite{Scene-LLM}, and LLaVA-3D~\cite{LLaVA-3D} leverage camera parameters to project multi-view 2D features into corresponding 3D coordinates, forming spatially consistent representations. GPT4Scene~\cite{GPT4Scene} and Struct2D~\cite{Struct2D} introduce bird’s-eye-view (BEV) representations to capture global scene layouts, while SpatialPIN~\cite{SpatialPIN}, MM-Spatial~\cite{MM-Spatial}, and GSReasoner~\cite{GSReasoner} employ depth maps to provide crucial depth cues. To capture spatio-temporal dynamics, See\&Trek~\cite{See_Trek} explicitly encodes egocentric trajectory maps, aiding camera-motion understanding during video capture. Furthermore, 3D foundation models such as VGGT~\cite{VGGT} and CUT3R~\cite{CUT3R} are integrated or distilled into MLLMs, for example, in Spatial-MLLM~\cite{Spatial-MLLM}, VLM-3R~\cite{VLM-3R}, and 3DThinker~\cite{3DThinker}, to extract 3D-reconstructive tokens from 2D imagery.

Vanilla MLLM–based approaches aim to improve spatial reasoning without incorporating explicit 3D priors. SpatialVLM~\cite{SpatialVLM} leverages large-scale scene-centric datasets to enhance spatial awareness, while Video3DLLM~\cite{Video3DLLM} extends this idea to multi-frame scenarios. SpaceR~\cite{SpaceR} introduces 2D grids with object-layout intermediate supervision to guide learning, and ST-Think~\cite{ST-Think} integrates reverse reasoning into reinforcement learning to improve spatial inference. R1-Zero-VSI~\cite{R1-Zero-VSI} constructs a high-quality spatial reasoning dataset and fine-tunes MLLMs using an optimized GRPO algorithm, whereas Spatial-Ladder~\cite{Spatial-Ladder} adopts a three-stage training strategy to progressively enhance spatial understanding. However, these methods depend on additional supervision or large-scale data, resulting in substantial training costs and limited generalization.

\section{Methodology}
\label{sec:framework}

\subsection{Formulation}

Given a sequence of $N$ temporally ordered frames $\mathcal{I}=\{I_1, I_2, \dots, I_N\}$ sampled from a video depicting a scene, and a natural language question $Q$, the objective is to predict the corresponding answer $A$ using an MLLM:
\begin{equation}
    A = \mathcal{F}_{\theta}(\mathcal{I}, Q)
\end{equation}
\noindent where $\mathcal{F}_{\theta}$ denotes an MLLM parameterized by $\theta$.

In contrast to single-frame visual reasoning, answering $Q$ from multi-frame observations requires the model to infer a coherent spatial context from partial views acquired across different viewpoints and time steps, while simultaneously constructing a task-relevant reasoning structure.

\noindent \textbf{Global Context.} Let $\mathcal{O}_i$ and $\mathcal{R}_i$ denote the sets of objects and intra-frame spatial relations observed in frame $I_i$, respectively. Each frame induces a local relational graph $\mathcal{G}_i=(\mathcal{O}_i,\mathcal{R}_i)$. The task requires integrating these partial observations into a global scene context $\mathcal{G}_{\mathrm{ctx}}=(\mathcal{O},\mathcal{R})$, where $\mathcal{O}=\bigcup_i \mathcal{O}_i$, and $\mathcal{R}$ includes both intra-frame relations and cross-frame relations established through object correspondences and viewpoint transitions.

\noindent \textbf{Task-Relevant Context.} To support question-oriented reasoning, we identify the spatial context relevant to $Q$. Specifically, the question-relevant object set is defined as $\mathcal{O}_{\mathrm{rel}}=\mathcal{O}_{\mathrm{exp}}\cup\mathcal{O}_{\mathrm{imp}}$, where $\mathcal{O}_{\mathrm{exp}}$ denotes objects explicitly mentioned in $Q$, and $\mathcal{O}_{\mathrm{imp}}$ denotes implicit objects serving as intermediate spatial anchors for multi-step reasoning. Let $\mathcal{R}_{\mathrm{rel}}\subseteq \mathcal{R}$ denote the corresponding question-relevant spatial relations. The resulting question-relevant context is defined as $\mathcal{G}_{\mathrm{rel}}=(\mathcal{O}_{\mathrm{rel}},\mathcal{R}_{\mathrm{rel}})$.

Therefore, accurate multi-frame spatial reasoning requires the model not only to construct a coherent global scene graph from distributed observations, but also to retrieve and reason over the question-relevant subgraph. By integrating global spatial context for cross-frame scene understanding with task-oriented context for question-specific reasoning, the MLLM can establish the spatial associations necessary for predicting the correct answer.

\begin{figure*}[t]
  \centering
  \includegraphics[width=1\linewidth]{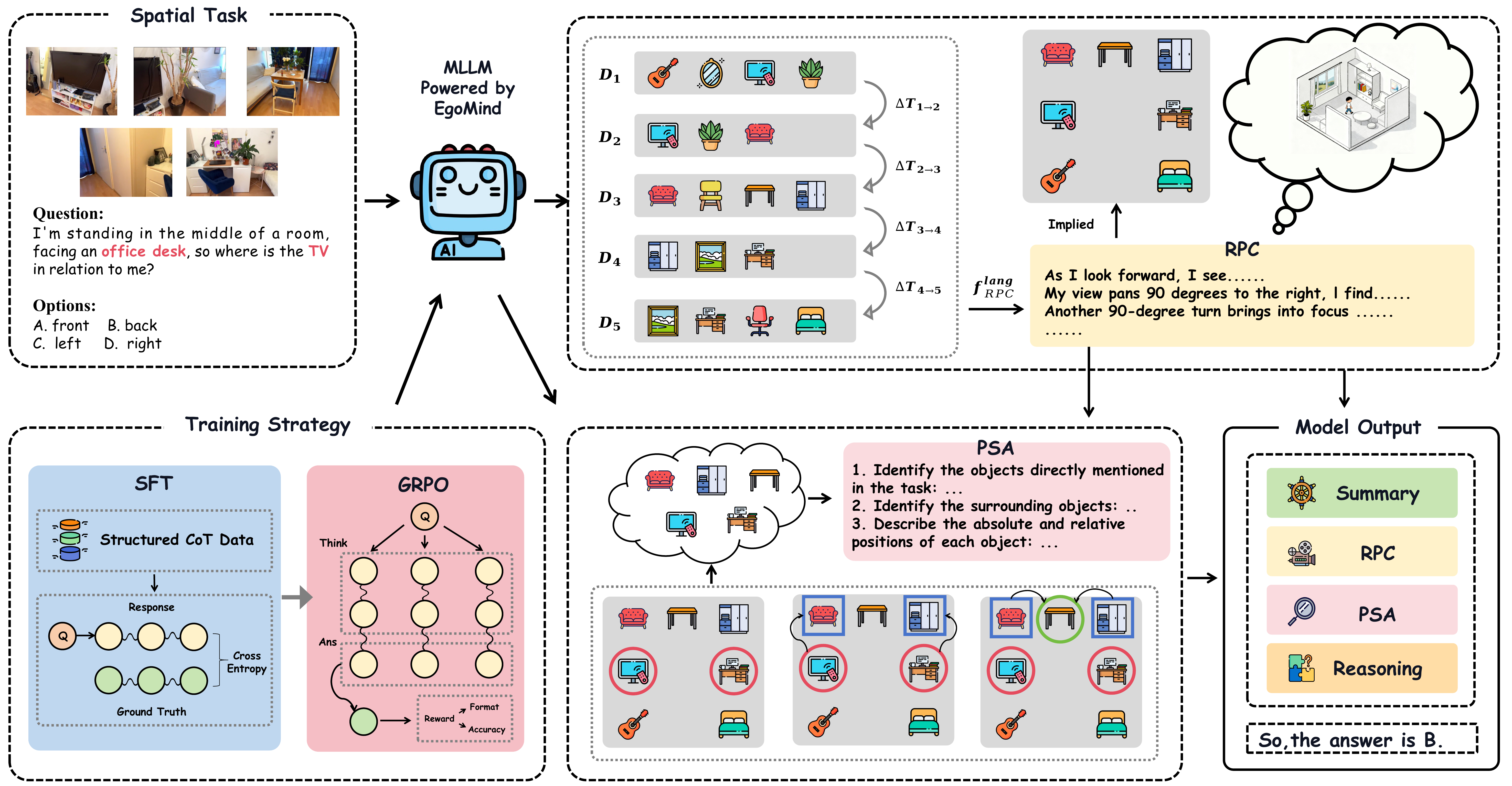}
   \caption{Illustration of the proposed EgoMind framework. MLLMs powered by EgoMind first generate a Role-Play Caption by producing per-frame scene descriptions and inferring viewpoint transitions. The model then performs Progressive Spatial Analysis (PSA) to identify relevant objects, expand spatial dependencies via implicit spatial bridges, and form a coherent reasoning chain. Finally, the system outputs the EgoMind CoT, unifying RPC and PSA into an interpretable spatial reasoning process.}
   \label{fig:framework}
\end{figure*}

\subsection{Role-Play Caption}
\label{sec:rpc}

To accurately answer spatial questions, most existing approaches~\cite{GPT4Scene, Struct2D} introduce additional 3D inputs to provide explicit geometric priors, thereby guiding MLLMs to construct the global spatial context $\mathcal{G}_{\mathrm{ctx}}$. Other approaches focus on predicting per-frame objects $\mathcal{O}_i$~\cite{Spatial-Mind} or estimating inter-frame camera motion, i.e., the pose transformation $\mathcal{V}_{i \rightarrow i+1}$~\cite{See_Trek}. However, such methods often struggle to establish reliable inter-frame relations $\mathcal{R}_{\mathrm{inter}}$, resulting in fragmented and spatially inconsistent scene understanding.

In contrast, EgoMind aims to construct the global spatial context $\mathcal{G}_{\mathrm{ctx}}$ purely through linguistic reasoning, without relying on explicit 3D priors. To form a coherent and cross-frame consistent spatial graph, two key aspects must be addressed. First, viewpoint transitions across frames should be explicitly captured to ensure continuity and spatial consistency. Second, anchor objects must be identified to connect overlapping observations across frames, thereby establishing a unified global representation.

To this end, we first derive a linguistic description $\mathcal{D}_i$ for each frame $I_i$. Each description encapsulates the detected objects $\mathcal{O}_i$ and their spatial configuration, enabling the model to reason about spatial layout and viewpoint transitions in purely linguistic form. The collection of frame-level descriptions forms the base context 
$\{\mathcal{D}_1, \mathcal{D}_2, \dots, \mathcal{D}_N\}$,
which serves as the structured linguistic input for subsequent reasoning.

Building on these frame-level descriptions, RPC further introduces transition descriptions $\Delta \mathcal{T}_{i \rightarrow i+1}$ between consecutive frames $\mathcal{D}_i$ and $\mathcal{D}_{i+1}$ to explicitly model viewpoint transitions from a first-person egocentric perspective. Each transition $\Delta \mathcal{T}_{i \rightarrow i+1}$ linguistically approximates the unobserved relative motion $\mathcal{V}_{i \rightarrow i+1}$ (e.g., \textit{“I move forward and turn right to view the table from another side”}), allowing the model to align frame-level observations coherently in space. Formally, the enriched Role-Play Caption context is defined as:
\begin{equation}
    \hat{\mathcal{C}} = \{\mathcal{D}_1, \Delta \mathcal{T}_{1 \rightarrow 2}, \mathcal{D}_2, \Delta \mathcal{T}_{2 \rightarrow 3}, \dots, \mathcal{D}_N\}
\end{equation}

To maintain narrative coherence, redundant object descriptions across adjacent frames are simplified through perspective normalization, such that each newly observed object or relation is incrementally integrated into the evolving scene context. This process yields a coherent, linguistically grounded scene representation that implicitly encodes both inter-frame correspondences and spatial continuity:
\begin{equation}
\hat{\mathcal{G}}_{\mathrm{RPC}} = f_{\mathrm{RPC}}^{\mathrm{lang}}(\hat{\mathcal{C}}) = (\hat{\mathcal{O}}, \hat{\mathcal{R}}, \hat{\mathcal{V}})
\end{equation}
\noindent where $\hat{\mathcal{O}}$ and $\hat{\mathcal{R}}$ denote the linguistically reconstructed objects and spatial relations, and $\hat{\mathcal{V}}$ denotes the inferred viewpoint transitions. Here, $f_{\mathrm{RPC}}^{\mathrm{lang}}(\cdot)$ represents the linguistic reasoning function performed by the model.

The resulting $\hat{\mathcal{G}}_{\mathrm{RPC}}$ provides a unified linguistic spatial graph that captures both intra-frame and inter-frame dependencies, serving as the foundation for higher-level spatial reasoning in EgoMind.

\subsection{Progressive Spatial Analysis}
\label{sec:psa}
Previous attempts aim to directly extract the question-relevant objects $\mathcal{O}_{\mathrm{rel}}$ and their relations from the global context. However, due to inaccurate object grounding and incomplete inter-frame associations, such direct inference is often affected by missing or noisy objects and relations, resulting in suboptimal reasoning chains and answers.

In contrast, we propose PSA as a key component of the EgoMind CoT for capturing task-relevant context. Given a question $Q$, PSA first identifies the explicitly mentioned target object set $\mathcal{O}_{\mathrm{exp}} = \{o_1, o_2, \dots, o_k\}$. It then expands this initial set by iteratively exploring the corresponding spatial neighborhoods in the linguistic scene graph $\hat{\mathcal{G}}_{\mathrm{RPC}}$ constructed by RPC. Finally, the model evaluates the spatial relations among the resulting consolidated objects.

Formally, for each explicit target object $o_i \in \mathcal{O}_{\mathrm{exp}}$, its spatial neighborhood, including itself, is defined as

\begin{equation}
    \mathcal{N}(o_i) = \left\{ o_j \in \hat{\mathcal{O}} \mid (o_i, o_j) \in \hat{\mathcal{R}} \right\}
\end{equation}

To ensure comprehensive coverage of potential spatial bridges, PSA aggregates these neighborhoods to form an expanded candidate set
$\hat{\mathcal{O}}_{\mathrm{rel}} = \bigcup_{o_i \in \mathcal{O}_{\mathrm{exp}}} \mathcal{N}(o_i)$.
This aggregated set $\hat{\mathcal{O}}_{\mathrm{rel}}$ is intended to cover the question-relevant objects $\mathcal{O}_{\mathrm{rel}}$, including both the explicitly mentioned targets $\mathcal{O}_{\mathrm{exp}}$ and the implicit spatial anchors $\mathcal{O}_{\mathrm{imp}}$.

Based on the expanded candidate set $\hat{\mathcal{O}}_{\mathrm{rel}}$, PSA further constructs a localized reasoning chain by exploring relational paths within the global context $\hat{\mathcal{G}}_{\mathrm{RPC}}$. Each step corresponds to an atomic spatial relation in $\hat{\mathcal{R}}$, ultimately yielding the task-relevant relation set $\hat{\mathcal{R}}_{\mathrm{rel}}$. The resulting reasoning process is formalized as:
\begin{equation}
    \hat{\mathcal{G}}_{\mathrm{PSA}} = f_{\mathrm{PSA}}^{\mathrm{lang}}(Q, \hat{\mathcal{G}}_{\mathrm{RPC}}) = 
    (
        \hat{\mathcal{O}}_{\mathrm{rel}},
        \hat{\mathcal{R}}_{\mathrm{rel}}
    )
\end{equation}

\noindent where $f_{\mathrm{PSA}}^{\mathrm{lang}}(\cdot)$ denotes the linguistic reasoning function. This function leverages the global scene graph $\hat{\mathcal{G}}_{\mathrm{RPC}}$ to derive a task-specific reasoning context $\hat{\mathcal{G}}_{\mathrm{PSA}}$.

By progressively expanding the reasoning scope and leveraging spatial bridges, PSA enables the model to perform fine-grained spatial reasoning without relying on explicit 3D geometry, thereby complementing the global scene graph constructed by RPC.

\subsection{Framework}
\label{sec:overall_framework}
\textbf{CoT Design.} Building on RPC for global context construction and PSA for task-relevant context extraction, we formulate the final chain-of-thought (CoT) structure of EgoMind, as illustrated in Fig.~\ref{fig:framework}.

The CoT begins with a \textit{Summary Field}, in which the model analyzes the question $Q$ to identify its spatial reasoning requirements and outline a high-level reasoning plan. Next, the \textit{RPC Field} generates detailed language-based scene descriptions, constructing a linguistically grounded global spatial context $\hat{\mathcal{G}}_{\mathrm{RPC}}$ that captures both intra-frame and inter-frame relations. The \textit{PSA Field} then progressively aggregates question-relevant objects and their spatial relations to derive a task-specific spatial context $\hat{\mathcal{G}}_{\mathrm{PSA}}$. Finally, the \textit{Reasoning Field} integrates the contextual information derived from the previous stages to produce the answer.

Since the underlying 3D scene context is inherently unobservable from discrete 2D frames, the constructed $\hat{\mathcal{G}}_{\mathrm{RPC}}$ and $\hat{\mathcal{G}}_{\mathrm{PSA}}$ serve as explicit linguistic context for spatial reasoning. Accordingly, the final inference process is formulated as:
\begin{equation}
    A = \mathcal{F}_{\theta}(\mathcal{I}, Q \mid \hat{\mathcal{G}}_{\mathrm{RPC}}, \hat{\mathcal{G}}_{\mathrm{PSA}})
\end{equation}
\noindent where $\mathcal{F}_{\theta}$ leverages both the global spatial context $\hat{\mathcal{G}}_{\mathrm{RPC}}$ and the task-specific spatial context $\hat{\mathcal{G}}_{\mathrm{PSA}}$ to derive the answer through a coherent and interpretable reasoning chain. By following this paradigm, MLLMs can systematically align multi-frame observations and perform fine-grained spatial reasoning, thereby achieving robust multi-view spatial understanding.

\begin{figure*}[!ht]
  \centering
  \includegraphics[width=1\linewidth]{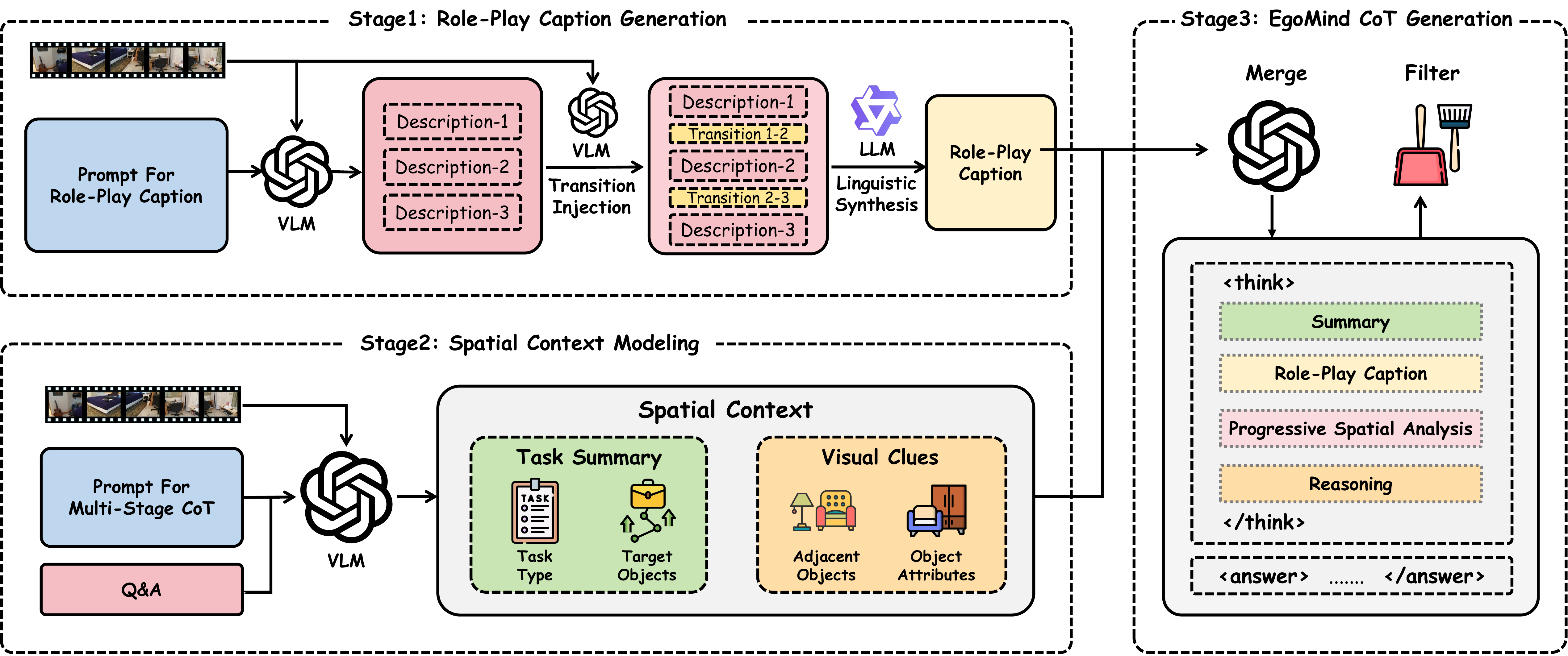}
   \caption{Illustration of the data generation pipeline. Randomly sampled video frames and a tailored instruction are first given to GPT-4o to produce detailed per-frame descriptions and infer viewpoint transitions between adjacent frames. Qwen2.5-72B then synthesizes these descriptions and transitions into the Role-Play Caption (RPC). In parallel, another GPT-4o instance, guided by a structured instruction, extracts the required spatial context from the multi-frame input and question. Finally, GPT-4o merges the RPC and spatial context to generate the final EgoMind Chain-of-Thought.}
   
   \label{fig:pipeline}
   \vspace{-0.4cm}
\end{figure*}

\noindent \textbf{Data Generation.} To enable MLLMs to follow the proposed CoT design, we develop a fully automated pipeline for generating EgoMind-style CoT data, as shown in Fig.~\ref{fig:pipeline}.

\textit{RPC Generation.} We first feed sampled multi-frame inputs into GPT-4o to generate frame-level descriptions $\mathcal{D}_i$. To avoid question-induced attention bias, we use a prompt that encourages exhaustive and unbiased descriptions. GPT-4o is then used to infer the viewpoint transition $\Delta \mathcal{T}_{i \rightarrow i+1}$ between adjacent frames. Based on these descriptions and transitions, we employ Qwen2.5-72B as $f_{\mathrm{RPC}}^{\mathrm{lang}}$ to produce linguistically grounded representations that encode inter-frame correspondences and spatial continuity.

\textit{Spatial Context Modeling.} To support PSA, we next construct task-relevant spatial context. Treating this as a pure VLM task, we provide GPT-4o with the sampled frames, the question, and a structured prompt that instructs it to generate a \textit{task summary} identifying the task type and target objects, as well as \textit{visual clues} describing adjacent objects and the attributes of target and neighboring entities.

\textit{EgoMind CoT Generation.} Finally, the generated RPC and extracted spatial context are fed into GPT-4o, which produces the full EgoMind CoT through a designed prompt template, integrating the summary, RPC, PSA, and reasoning stages.

Unlike existing approaches that rely on large-scale multimodal data collection, manual annotation, or structured geometric labels, our pipeline is entirely annotation-free and highly scalable. Using this pipeline, we generate 5K high-quality CoT samples to substantially enhance the spatial reasoning capability of MLLMs.

\noindent \textbf{Training Strategy.}
To train MLLMs to follow the EgoMind CoT structure, we adopt a two-stage paradigm: Supervised Fine-Tuning (SFT) to learn the structured CoT format, followed by Group Relative Policy Optimization (GRPO) to further improve reasoning through reward-guided refinement.

For each question $q$, GRPO samples a group of candidate reasoning paths $\{o_1,\dots,o_G\}$ from the old policy $\pi_{\theta_{\mathrm{old}}}$ and optimizes:
\begin{equation}
\small
\begin{aligned}
\mathcal{J}_{\mathrm{GRPO}}(\theta)
= &\mathbb{E}_{q,\,o_i}\Bigg[
\frac{1}{G}\sum_{i=1}^{G}
\Bigg(
\min \Big(
r_i(\theta)A_i,
\\
&\operatorname{clip}(r_i(\theta),1-\varepsilon,1+\varepsilon)A_i
\Big)
-\beta\,\operatorname{KL}\!\left(
\pi_\theta \,\|\, \pi_{\mathrm{ref}}
\right)
\Bigg)
\Bigg]
\end{aligned}
\label{eq:grpo_multiline}
\end{equation}
where
\[
r_i(\theta)=\frac{\pi_\theta(o_i\mid q)}{\pi_{\theta_{\mathrm{old}}}(o_i\mid q)}
\]
is the importance-sampling ratio, and
\[
A_i=\frac{R_i-\operatorname{mean}(R_1,\dots,R_G)}{\operatorname{std}(R_1,\dots,R_G)}
\]
is the group-normalized advantage. The KL term regularizes $\pi_\theta$ toward the reference policy $\pi_{\mathrm{ref}}$, with strength controlled by $\beta$. The reward is defined as:
\begin{equation}R_i=w_f\,R_{\mathrm{format}}(o_i\mid q)+w_a\,R_{\mathrm{accuracy}}(o_i\mid q)\end{equation}
where $w_f$ and $w_a$ balance format and accuracy rewards.

Based on these reward signals, GRPO iteratively refines the model’s policy to improve both structural adherence and answer accuracy, ultimately enabling the MLLM to internalize the EgoMind reasoning paradigm.

\begin{table*}[!tbp]
    \caption{A comprehensive comparison of EgoMind with state-of-the-art vision–language models across four spatial reasoning benchmarks.}
    \label{tab:model_benchmark_simplified}
    \centering
    \resizebox{\textwidth}{!}{
    \begin{tabular}{l c c c cccc cccc c c c}
        \toprule
        \multirow{3}{*}{Models} & \multirow{3}{*}{Params} & \multirow{3}{*}{Data Size} & 
        \multicolumn{9}{c}{VSI-Bench} & 
        \multirow{3}{*}{SPAR-Bench} & \multirow{3}{*}{SITE-Bench} & \multirow{3}{*}{SPBench} \\
        \cmidrule(lr){4-12}
        & & & \multirow{2}{*}{Overall} & \multicolumn{4}{c}{Numerical Question} & 
        \multicolumn{4}{c}{Multiple-Choice Question} & & & \\
        \cmidrule(lr){5-8} \cmidrule(lr){9-12}
        & & & & Obj. Cnt. & Abs. Dist. & Obj. Size & Room Size & 
        Rel. Dist. & Rel. Dir. & Route Plan & Appr. Order & & & \\
        \midrule
        
        \multicolumn{15}{l}{\textit{Closed-source Models}} \\
        \midrule
        GPT-4.1 & - & - & 47.21 & 44.44 & 26.16 & 64.71 & 53.06 & 51.27 & 36.47 & 38.14 & 63.43 & 42.77 & 64.16 & 54.51 \\
        GPT-5 & - & - & 51.66 & 45.82 & 32.57 & 69.77 & 46.15 & 57.61 & 41.74 & 47.42 & 67.48 & 54.72 & 67.89 & 54.77 \\
        Gemini 2.5 Pro & - & - & 50.62 & 40.8 & 35.13 & 67.00 & 54.83 & 55.92 & 41.74 & 45.36 & 62.78 & 49.42 & 66.00 & 55.90 \\
        \midrule

        \multicolumn{15}{l}{\textit{Pure Image--Language Models}} \\
        \midrule
        Qwen2.5-VL~\cite{Qwen25VL} & 7B & - & 30.02 & 23.75 & 10.50 & 36.98 & 35.31 & 38.45 & 37.09 & 28.87 & 28.48 & 33.19 & 53.74 & 41.65 \\
        InternVL3.5~\cite{InternVL3.5} & 7B & - & 44.17 & 50.37 & 34.66 & 50.89 & 46.91 & 45.35 & 46.18 & 34.54 & 38.19 & 38.05 & 52.09 & 52.23 \\
        MiMo-VL-RL~\cite{MIMOVL} & 7B & - & 33.28 & 16.90 & 20.41 & 45.88 & 30.76 & 41.27 & 28.41 & 31.44 & 46.44 & 31.44 & 40.87 & 33.30 \\
        Ovis2.5~\cite{Ovis2.5} & 7B & - & 40.93 & 50.60 & 32.60 & 54.92 & 37.71 & 43.10 & 34.81 & 28.87 & 34.14 & 51.44 & 59.35 & 25.75 \\
        MiniCPM-V 4.5~\cite{MINICPM-V4.5} & 7B & - & 34.13 & 42.41 & 25.66 & 32.55 & 20.21 & 40.42 & 36.57 & 22.68 & 39.48 & 35.98 & 55.93 & 38.61 \\
        LLaVA-OneVision~\cite{LLaVA}& 7B & - & 32.40 & 47.70 & 20.20 & 47.40 & 12.30 & 42.50 & 35.20 & 29.40 & 24.40 & 30.60 & - & 32.70 \\
        \midrule
        
        \multicolumn{15}{l}{\textit{Models with Explicit 3D Spatial Priors (Point Clouds, Depth, BEV, Camera Pose, Trajectory)}} \\
        \midrule
        GPT4Scene~\cite{GPT4Scene} & 7B & - & 24.87 & 44.04 & 2.4 & 27.12 & 25.94 & 43.10 & 16.01 & 31.96 & 24.43 & 26.53 & 50.04 & 23.52 \\
        Struct-2D~\cite{Struct2D} & 7B & 200k & 43.60 & - & - & - & - & - & - & - & - & - & - & - \\
        SpaceVista~\cite{SpaceVista} & 7B & 1M & 48.60 & - & - & - & - & - & - & - & - & 41.60 & - & - \\
        SEE\&TREK~\cite{See_Trek}& 14B & - & 45.60 & 65.90 & 35.70 & 50.50 & 48.40 & 49.00 & 41.00 & 27.80 & 46.80 & - & - & - \\
        Spatial-MLLM~\cite{Spatial-MLLM}& 4B & 120K & 48.40 & 65.30 & 34.80 & 63.10 & 45.10 & 41.30 & 46.20 & 33.50 & 46.30 &  35.10 & 43.99 & 48.40 \\
        \midrule
        
        \multicolumn{15}{l}{\textit{Models without Explicit 3D Inputs}} \\
        \midrule
        SpaceR \cite{SpaceR} & 7B & 151k & 45.76 & 57.10 & 30.07 & 60.75 & 32.67 & 43.80 & 45.56 & 30.41 & 46.93 & \underline{38.26} & \underline{56.48} & \underline{53.39} \\
        Video-R1 \cite{Video-R1}& 7B & 260k & 37.10 & - & - & - & - & - & - & - & - & - & - & - \\
        Spatial-Mind \cite{Spatial-Mind}& 7B & 925K & 43.90 & 55.00 & 29.50 & 57.30 & 44.00 & 43.50 & 44.30 & 38.30 & 39.20 & - & - & - \\
        R1-Zero-VSI \cite{R1-Zero-VSI}& 7B & 75k & 40.70 & 59.90 & 29.60 & 50.80 & 48.30 & 35.40 & 35.60 & 34.00 & 31.50 & - & - & - \\
        ViLaSR \cite{VILASR}& 7B & 80k & \underline{46.35} & 56.58 & 33.67 & 58.57 & 30.49 & 42.11 & 45.76 & 25.26 & 55.02 & 34.98 & 56.09 & 53.16 \\
        \textbf{EgoMind} (\textit{Ours}) & 7B & 25k & \textbf{50.16} & 54.51 & 37.94 & 67.12 & 40.35 & 44.08 & 47.21 & 31.96 & 58.41 & \textbf{39.03} & \textbf{58.03} & \textbf{55.02} \\
        \midrule
        
        \bottomrule
    \end{tabular}
    }
\end{table*}

\section{Experiments}
\label{sec:experiments}

\subsection{Implementation}

Following prior works~\cite{SpaceR, Video-R1}, we adopt Qwen2.5-VL-7B~\cite{Qwen25VL} as our base model for a fair comparison. To enhance spatial reasoning, we follow the two-stage training strategy described in Sec.~\ref{sec:overall_framework}. Specifically, supervised fine-tuning (SFT) is performed on 5K automatically generated EgoMind samples using the LLaMA-Factory framework~\cite{LLaMAFactory} for 3 epochs with a learning rate of $5 \times 10^{-6}$.

For reinforcement learning, we randomly sample 20K examples from SpaceR-91k~\cite{SpaceR} and conduct GRPO training using the EasyR1 framework~\cite{EasyR1}. During this stage, we use a batch size of 64 and generate 8 candidate reasoning paths for each question. For both training and inference, we uniformly sample 16 frames from each video. The visual input resolution is capped at $256 \times 28 \times 28$, and inputs exceeding this limit are proportionally downsampled. Additional implementation details are provided in the Appendix.

We evaluate EgoMind on four spatial reasoning benchmarks: VSI-Bench~\cite{VSI-Bench}, SITE-Bench~\cite{SITE-Bench}, SPAR-Bench~\cite{SPAR-Bench}, and SPBench~\cite{Spatial-Ladder}. These benchmarks include both multiple-choice and numerical reasoning tasks, covering abilities such as spatial memory, cross-view understanding, and global scene consistency. For multiple-choice questions, we report \textit{Accuracy (ACC)} based on exact match with ground truth. For numerical questions, we use \textit{Mean Relative Accuracy (MRA)}~\cite{VSI-Bench} as the evaluation metric.

\subsection{Comparison with State-of-the-Art}

We compare EgoMind against four categories of models: (i) closed-source models, (ii) pure image--language models, (iii) models with explicit 3D spatial priors, and (iv) models without explicit 3D inputs. Evaluations are conducted on VSI-Bench, SPAR-Bench, SITE-Bench, and SPBench, and the results are summarized in Tab.~\ref{tab:model_benchmark_simplified}.

As shown in Tab.~\ref{tab:model_benchmark_simplified}, EgoMind achieves highly competitive performance across all four benchmarks. Compared with Qwen2.5-VL-7B, its base model, EgoMind improves performance from 30.02 to 50.16 on VSI-Bench and from 41.65 to 55.02 on SPBench using only 25K training samples (5K CoT-supervised and 20K RL samples). These gains indicate that the proposed CoT formulation effectively unlocks the latent spatial reasoning capabilities of MLLMs, enabling strong multi-frame spatial understanding without additional modalities.

Relative to models trained without explicit 3D inputs, EgoMind consistently delivers superior or comparable performance, despite these baselines relying on substantially larger training sets and diverse forms of spatial or geometric supervision. In particular, EgoMind outperforms ViLaSR, which is trained on 80K spatially annotated samples, across all benchmarks, highlighting the data efficiency and strong generalization of our linguistic reasoning paradigm. These results further validate the effectiveness of the EgoMind CoT formulation in enabling spatial reasoning that is both effective and data-efficient.

Comparisons with methods that incorporate explicit 3D spatial priors further highlight EgoMind's effectiveness. Although the data volume used by EgoMind is only 2.5\% of the training data required by SpaceVista (1M samples), it achieves higher performance on VSI-Bench (50.16 vs.\ 48.60) and remains competitive on SPAR-Bench, demonstrating strong spatial generalization while avoiding the substantial overhead of 3D data alignment. More broadly, these findings suggest that the richer spatial context induced by linguistic reasoning may help improve the spatial generalization of MLLMs across diverse spatial cognition tasks, even without geometric priors or 3D supervision.

\begin{table}[htbp]
    \centering
    \caption{Ablation study of different components in EgoMind evaluated on VSI-Bench. RPC and PSA denote Role-Play Caption and Progressive Spatial Analysis, respectively.}
    \label{tab:ablation_study}
    \begin{tabular*}{0.9\linewidth}{@{\extracolsep{\fill}} l c c c c @{}}
        \toprule
        \multirow{2}{*}{\textbf{Exp}} & \multirow{2}{*}{\textbf{RPC}} & \multirow{2}{*}{\textbf{PSA}} & \multicolumn{2}{c}{\textbf{VSI-Bench}} \\
        \cmidrule(lr){4-5}
        & & & \textbf{+SFT} & \textbf{+RL} \\
        \midrule
        \textcolor{gray}{Baseline} & \textcolor{gray}{--} & \textcolor{gray}{--} & \multicolumn{2}{c}{\textcolor{gray}{30.02 {\scriptsize (vanilla)}}} \\
        \midrule 
        Full              & \checkmark  & \checkmark  & \textbf{42.33} & \textbf{50.16} \\
        w/o RPC           & $\times$    & \checkmark  & 41.52 & 47.69 \\
        w/o PSA           & \checkmark  & $\times$    & 41.23 & 45.15 \\
        \bottomrule
    \end{tabular*}
\end{table}

\subsection{Ablation Study}

\noindent \textbf{CoT Components.} We conduct ablation studies on RPC and PSA within the EgoMind CoT framework to evaluate their respective contributions. We assess the model on VSI-Bench at two stages: first, after SFT only, to isolate the direct gain brought by the CoT template itself; and second, after the full pipeline including RL, to evaluate the additional improvement achieved when the model is further trained to reason with the EgoMind CoT formulation.

As shown in Tab.~\ref{tab:ablation_study}, the full EgoMind CoT formulation yields substantial gains over the baseline. Under SFT alone, performance improves from 30.02 to 42.33, and the subsequent RL stage further raises it to 50.16, demonstrating the effectiveness of combining structured linguistic reasoning with RL for multi-frame spatial understanding. The contribution of the core components becomes even more pronounced after RL. Removing RPC causes only a modest drop during SFT (42.33 $\rightarrow$ 41.52), but leads to a much larger degradation after RL (50.16 $\rightarrow$ 47.69), confirming that modeling egocentric global context is critical for constructing a comprehensive spatial representation. Likewise, removing PSA reduces the SFT score to 41.23 and the RL score to 45.15, suggesting that PSA helps the model capture implicit spatial cues and extend the reasoning chain.

\begin{table}[htbp]
    \centering
    \caption{Ablation study on CoT modifications evaluated on VSI-Bench. MFC: Multi-Frame Caption. CVP: Camera View Prediction. DSA: Direct Spatial Analysis.}
    \label{tab:cot_modification}
    \begin{tabular*}{0.9\linewidth}{@{\extracolsep{\fill}} l c c @{}}
        \toprule
        \multirow{2}{*}{\textbf{CoT Modification}} & \multicolumn{2}{c}{\textbf{VSI-Bench}} \\
        \cmidrule{2-3}
        & \textbf{+SFT} & \textbf{+RL} \\
        \midrule
        Full CoT             & \textbf{42.33} & \textbf{50.16} \\
        RPC $\rightarrow$ MFC          & 41.58 & 48.09 \\
        RPC $\rightarrow$ MFC + CVP    & 41.84 & 47.12 \\
        PSA $\rightarrow$ DSA          & 41.54 & 47.24 \\
        \bottomrule
    \end{tabular*}
\end{table}

\noindent \textbf{Candidate Variants.} To further investigate the design choices of our CoT framework, we conduct ablation studies on alternative component designs. For RPC, which serves as a global scene modeling module, we introduce a \textit{Multi-Frame Caption} (MFC) baseline that directly concatenates per-frame captions without modeling transitions, as well as an MFC variant augmented with \textit{Camera View Prediction} (CVP), which explicitly predicts numerical viewpoint transformations between frames. For PSA, which serves as a task-oriented reasoning module, we introduce a \textit{Direct Spatial Analysis} (DSA) variant that identifies all task-relevant objects at once rather than progressively expanding the search space. Comparisons are conducted on VSI-Bench after both the SFT and RL phases, and the results are reported in Tab.~\ref{tab:cot_modification}.

As shown in Tab.~\ref{tab:cot_modification}, replacing RPC with MFC lowers RL performance from 50.16 to 48.09, since MFC simply concatenates independent descriptions without modeling transitions, leading to weaker cross-frame coherence. Adding CVP slightly improves SFT but further reduces RL performance to 47.12, suggesting that noisy geometric predictions can mislead the reasoning policy. This highlights the robustness of RPC's language-driven transition modeling. Replacing PSA with DSA also degrades performance, reducing the SFT and RL scores to 41.54 and 47.24, respectively. Unlike PSA, DSA focuses only on explicit objects and local relations, lacking the progressive expansion needed to identify implicit spatial bridges and broader relational context.

\begin{table}[t]
\centering
\caption{Ablation on the number of input frames for Room Size Estimation on VSI-Bench, using Qwen2.5-VL-7B as the baseline.}
\small
\small
\begin{tabular}{lccccc}
\toprule
\# Frames & Baseline & 4 & 8 & 12 & 16 \\
\midrule
Score & 35.31 & 36.14 & 36.59 & 38.88 & 40.35 \\
Gain  & --    & +0.83 & +1.28 & +3.57 & +5.04 \\
\bottomrule
\end{tabular}
\label{tab:frame_ablation_roomsize}
\end{table}

\noindent \textbf{Discussion.} Although EgoMind primarily improves spatial reasoning through object- and relation-centric context modeling in RPC and PSA, we find that it also benefits metric-aware tasks, even without explicit geometric supervision. To better understand this effect, we ablate the number of RPC input frames in Tab.~\ref{tab:frame_ablation_roomsize}. As the table shows, performance on \textit{Room Size Estimation} improves consistently with more input frames. This suggests that EgoMind supports metric-aware reasoning by accumulating implicit scale and spatial continuity cues across views within a coherent global context. Hence, even without explicit 3D priors, linguistically structured cross-frame reasoning can facilitate scene-level metric understanding. More detailed case analysis is provided in the Appendix.

\section{Conclusion}
\label{sec:conclusion}

We present EgoMind, a CoT framework for geometry-free spatial reasoning in multimodal LLMs, built on RPC and PSA. By constructing a linguistic scene graph over multi-frame observations, EgoMind enables strong spatial cognition without 3D priors. Experiments across multiple benchmarks demonstrate its effectiveness, establishing it as a scalable, lightweight alternative to 3D-based methods.

\noindent \textbf{Limitations.} EgoMind is still constrained by limited temporal reasoning, insufficient synthetic trace diversity, and a lack of validation on larger MLLMs. Future work will focus on improving temporal consistency, enriching data diversity, and generalizing to long-horizon embodied tasks.

\section*{Acknowledgment}
This work is supported by the National Key Research and Development Plan (2024YFB3309300).

{
    \small
    \bibliographystyle{ieeenat_fullname}
    \bibliography{main}
}

\clearpage
\maketitlesupplementary

\appendix
\renewcommand{\thefigure}{\Alph{figure}}
\renewcommand{\thetable}{\Alph{table}}
\section{Implementation Details}
\label{sec:experiments_details}

\subsection{Training Strategy}
\label{sec:training_strategy}

Our training pipeline consists of two stages: Supervised Fine-Tuning (SFT) for initializing spatial reasoning ability and aligning the model with the EgoMind CoT format, followed by Reinforcement Learning (RL) to further enhance structured reasoning quality through GRPO.

\noindent \textbf{Supervised Fine-Tuning.} Based on the 5K automatically generated SFT samples described in Fig.~\ref{fig:pipeline} of the main paper, we fine-tune Qwen2.5-VL-7B using the LLaMA-Factory framework to provide the model with initial spatial reasoning ability and align its outputs with the EgoMind CoT format, forming a strong foundation for the subsequent RL stage. During SFT, 16 frames are uniformly sampled from each video, and the maximum pixel budget is constrained to 200{,}704 ($256 \times 28 \times 28$). Training is conducted for 3 epochs with a learning rate of $5 \times 10^{-6}$ using a cosine decay schedule and a warmup ratio of 0.1. We employ the AdamW optimizer and train in \texttt{bf16} precision to improve memory efficiency and stability. The LLM and projector components are set to be trainable, while the ViT backbone remains frozen.

To ensure consistency with the RL stage, we append a structured instruction prompt to each question, guiding the model to produce outputs that adhere to the EgoMind CoT format:
\definecolor{mybluebg}{HTML}{EAF6FC} 
\definecolor{myblueframe}{HTML}{2C7FB8} 
\begin{tcolorbox}[
   enhanced, 
   width = 0.95\linewidth,
   center,
    title = {Task Prompt},
    colback = mybluebg,
    colframe = myblueframe,
    coltitle = white, 
    colbacktitle = myblueframe, 
    fonttitle = \bfseries,
    rounded corners,
    arc = 2mm,
    boxrule = 1pt,
]
You should first think about the reasoning process in the mind and then provide
the user with the answer. The reasoning process and answer are enclosed within \texttt{<think>} \texttt{</think>} and \texttt{<answer>} \texttt{</answer>} tags, respectively, i.e., \texttt{<think>} reasoning process here \texttt{</think>} \texttt{<answer>} answer here \texttt{</answer> 
}

\end{tcolorbox}

\noindent \textbf{Reinforcement Learning.}
During the RL phase, we train the MLLM on 20K samples using the GRPO algorithm implemented in the EasyR1 framework. We set the batch size to 64, the learning rate to $1 \times 10^{-6}$, and apply a weight decay of $1.0 \times 10^{-2}$. The AdamW optimizer is adopted with \texttt{bf16} precision. To balance effective policy updates with controlled divergence from the reference model, we use a KL penalty coefficient of $1 \times 10^{-4}$. For each prompt, the policy generates 8 candidate reasoning paths to compute group-wise rewards, using a temperature of 1.0 and top-$p$ of 0.99. The maximum response length is capped at 2048 tokens. 

The total reward is defined as a weighted sum of a format reward ($R_{\text{format}}$) and an accuracy reward ($R_{\text{accuracy}}$), with weights 0.2 and 0.8, respectively. The format reward is binary: $R_{\text{format}} = 1$ if the model output strictly adheres to the required \texttt{think}-\texttt{answer} structure, and 0 otherwise. The accuracy reward $R_{\text{accuracy}}$ evaluates the content contained within the \texttt{<answer>} and \texttt{</answer>} tags. For multiple-choice questions, we assign a discrete score of 0 or 1 based on exact matching of the predicted option (A/B/C/D). For numerical questions, we compute accuracy using the Mean Relative Accuracy (MRA) metric, which measures the relative closeness between the predicted value and the ground truth.

\subsection{Inference Strategy}
\label{subsec:inference_strategy}

To ensure fair comparisons across models of varying architectures, we strictly standardize our evaluation protocol. For closed-source models (e.g., GPT-4.1, GPT-5, and Gemini 2.5 Pro), we apply the identical CoT prompt detailed in Section~\ref{sec:training_strategy}. For open-source MLLMs, we report the higher score between direct generation and CoT prompting to avoid penalizing models with weaker instruction-following capabilities. Furthermore, we evaluate all reasoning models under their official default configurations (e.g., ``dynamic thinking'' for Gemini 2.5 Pro and ``medium reasoning effort'' for GPT-5). 

Regarding visual inputs, we constrain the maximum image resolution to $256 \times 28 \times 28$ across all benchmarks to maintain strict consistency with our training phase. Additionally, for video-based tasks specifically, we uniformly sample 16 frames per video sequence.

\begin{figure*}[!htbp]
    \centering
    \includegraphics[width=\linewidth]{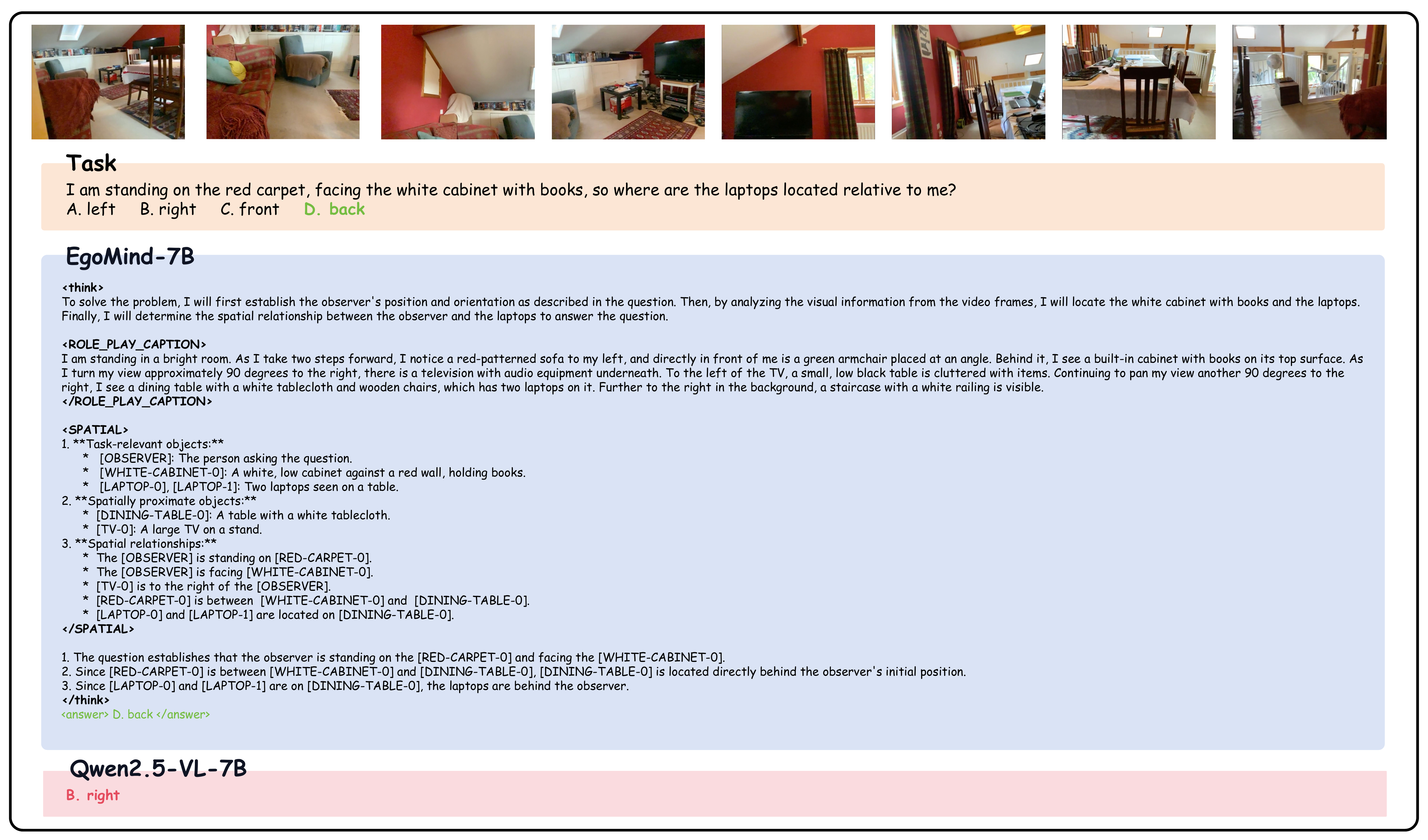}
    \caption{A case study of relational reasoning with the Qwen2.5-VL-7B model enhanced by the EgoMind framework.}
    \label{fig:qwen_case_en}
\end{figure*}

\subsection{Data Construction}

\paragraph{Supervised Fine-Tuning Data} For the SFT stage, our goal is to construct a compact yet diverse dataset that enables MLLMs to learn the EgoMind CoT format under strict cost constraints. To achieve this, we sample approximately 5K instances from the SpaceR-91k dataset. During sampling, we filter out trivial or overly ambiguous cases and enforce a more uniform distribution across different question types and answer patterns to reduce dataset-induced bias. The corresponding EgoMind CoT annotations are automatically generated using the pipeline described in Sec.~\ref{sec:overall_framework} of the main paper. Furthermore, to guarantee the reliability of the generated data, we employ Gemini 2.5 Pro driven by specifically tailored prompts to conduct comprehensive quality checks and filtering:

\begin{itemize}[leftmargin=1em] 
    \item \textbf{Hallucination Check:} We verify whether the finally merged chain-of-thought content factually conflicts with the input video frames, ensuring that no erroneous information is introduced during the Merge stage.
    
    \item \textbf{Logical Consistency:} We strictly examine the consistency between the PSA/RPC context and the Reasoning section. This guarantees that the reasoning conclusions are logically derived from the evidence provided by PSA and RPC.
    
    \item \textbf{Format \& Correctness:} We check whether the final extracted answer is correct by comparing it against the ground truth labels, and we ensure that the output format strictly complies with the training requirements.
\end{itemize}

\begin{table*}[!ht]
    \caption{Performance evaluation of Qwen2.5-VL models at different scales (3B and 7B) on the VSI-Bench benchmark. The table details the step-wise performance gains, comparing the base models with their counterparts enhanced through Supervised Fine-Tuning (SFT) and Reinforcement Learning (RL).}
    \label{tab:VSI-Bench_qwen_ablation}
    \centering
    \resizebox{\textwidth}{!}{
    \begin{tabular}{l l l c cccc cccc}
        \toprule
        \multirow{2}{*}{Models} & \multirow{2}{*}{Params} & \multirow{2}{*}{Version} & \multirow{2}{*}{Overall} & \multicolumn{4}{c}{Numerical Question} & \multicolumn{4}{c}{Multiple-Choice Question} \\
        \cmidrule(lr){5-8} \cmidrule(lr){9-12}
        & & & & Obj. Cnt. & Abs. Dist. & Obj. Size & Room Size & Rel. Dist. & Rel. Dir. & Route Plan & Appr. Order \\
        \midrule
        
        \multirow{3}{*}{Qwen2.5-VL-3B} & \multirow{3}{*}{3B} & Base & 27.68 & 27.06 & 19.78 & 25.37 & 19.24 & 32.54 & 39.26 & 28.87 & 22.33 \\
        &  & +SFT & 37.08 & 36.48 & 21.74 & 36.88 & 30.66 & 37.04 & 45.14 & 22.68 & 53.56 \\
        &  & +RL & 45.44 & 44.96 & 31.29 & 59.07 & 37.60 & 41.69 & 47.52 & 23.20 & 55.66 \\
        \midrule
        
        \multirow{3}{*}{Qwen2.5-VL-7B} & \multirow{3}{*}{7B} & Base & 30.02 & 23.75 & 10.50 & 36.98 & 35.31 & 38.45 & 37.09 & 28.87 & 28.48 \\
        &  & +SFT & 42.33 & 43.20 & 23.87 & 50.34 & 29.03 & 41.41 & 47.42 & 29.38 & 57.44 \\
        &  & +RL & 50.16 & 54.51 & 37.94 & 67.12 & 40.35 & 44.08 & 47.21 & 31.96 & 58.41 \\
        \bottomrule
    \end{tabular}
    }
\end{table*}

\paragraph{Reinforcement Learning Data} For the RL stage, we further sample 20K instances from the SpaceR-91k dataset. Since RL relies solely on outcome-based rewards and does not require CoT supervision, we remove extreme cases that are excessively easy or unsolvable. From the remaining pool, we select 20K moderately challenging samples to provide sufficient difficulty for policy improvement while maintaining stable reward signals.

\begin{figure*}[!htbp] 
    \centering
    \vspace{0.3cm}
    \includegraphics[width=\linewidth]{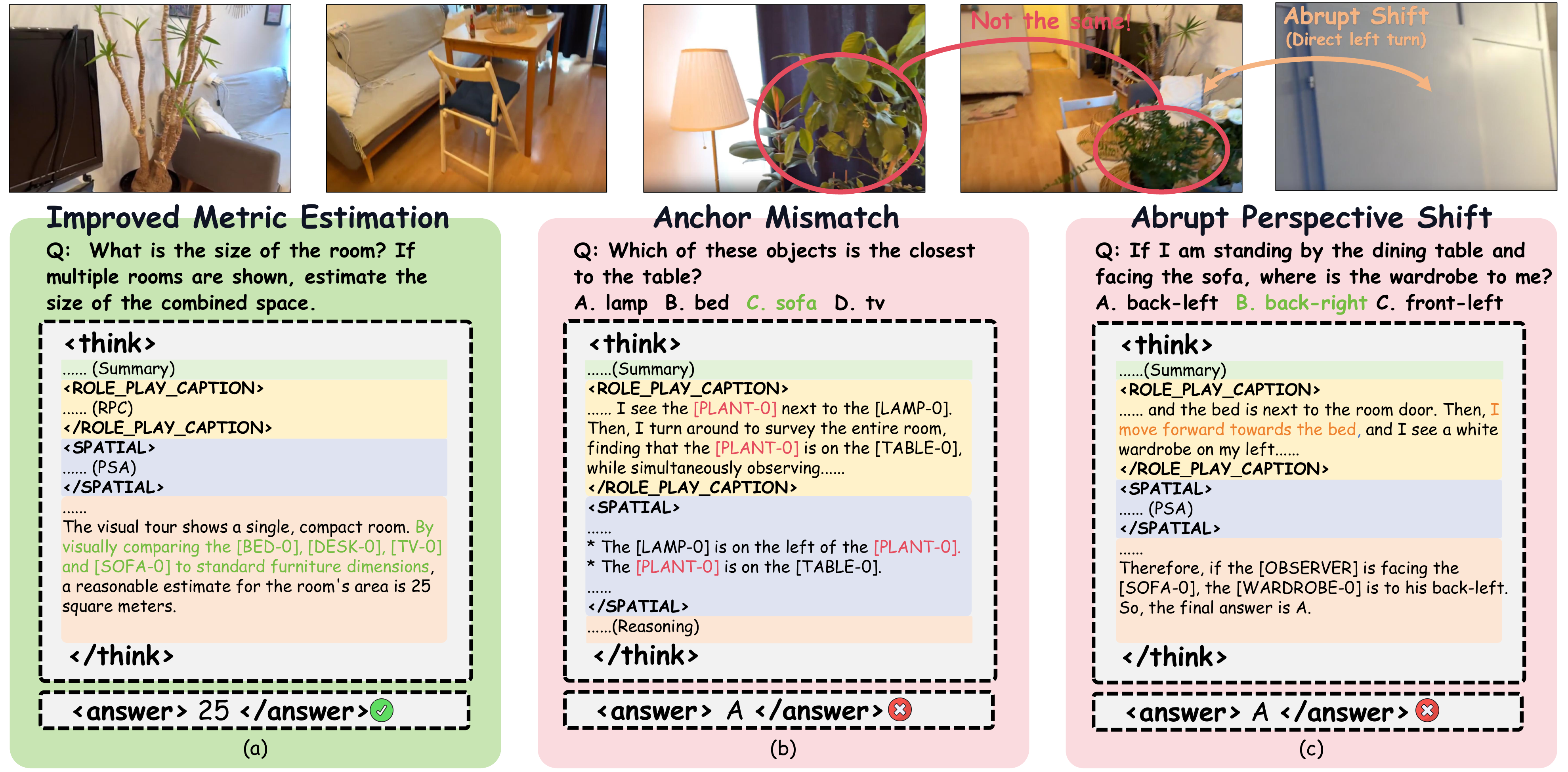}
    \caption{Case studies of EgoMind.}
    \label{fig:example}
\end{figure*}

\subsection{Benchmarks}
To comprehensively evaluate the spatial reasoning capability of EgoMind, we consider four representative benchmarks that cover a diverse set of spatial perception and reasoning tasks.

\noindent \textbf{VSI-Bench} assesses an MLLM’s ability to perceive, memorize, and reason about physical spaces through continuous visual observation. It contains 5,000 QA pairs across 288 real-world indoor videos sourced from ScanNet, ScanNet++, and ARKitScenes, and includes tasks such as object counting, distance estimation, relative direction prediction, and route planning.

\noindent \textbf{SPAR-Bench} provides over 7,200 human-verified QA samples spanning 20 spatial reasoning tasks, ranging from basic geometric perception to high-level relational reasoning. It uniquely employs only static images (single-view or multi-view), enabling a pure evaluation of a model’s ability to infer 3D spatial structure from discrete viewpoints without temporal information.

\noindent \textbf{SITE-Bench} integrates 30 existing spatial-intelligence datasets and augments them with newly designed tasks, offering a unified multiple-choice framework for systematic evaluation of spatial reasoning in MLLMs. In our experiments, we adopt the video-based subset of SITE-Bench, which contains 3,808 video QA tasks covering diverse spatial understanding scenarios.

\noindent \textbf{SPBench} comprises 1,328 QA pairs divided into two subsets: SPBench-SI (single-image, 1,009 QA) and SPBench-MV (multi-view, 319 QA). It is specifically designed to measure a model’s geometric understanding, object enumeration ability, and multi-view spatial synthesis. All samples are derived from indoor scenes in the ScanNet dataset.
\begin{figure*}[!htbp]
    \centering
    \includegraphics[width=\linewidth]{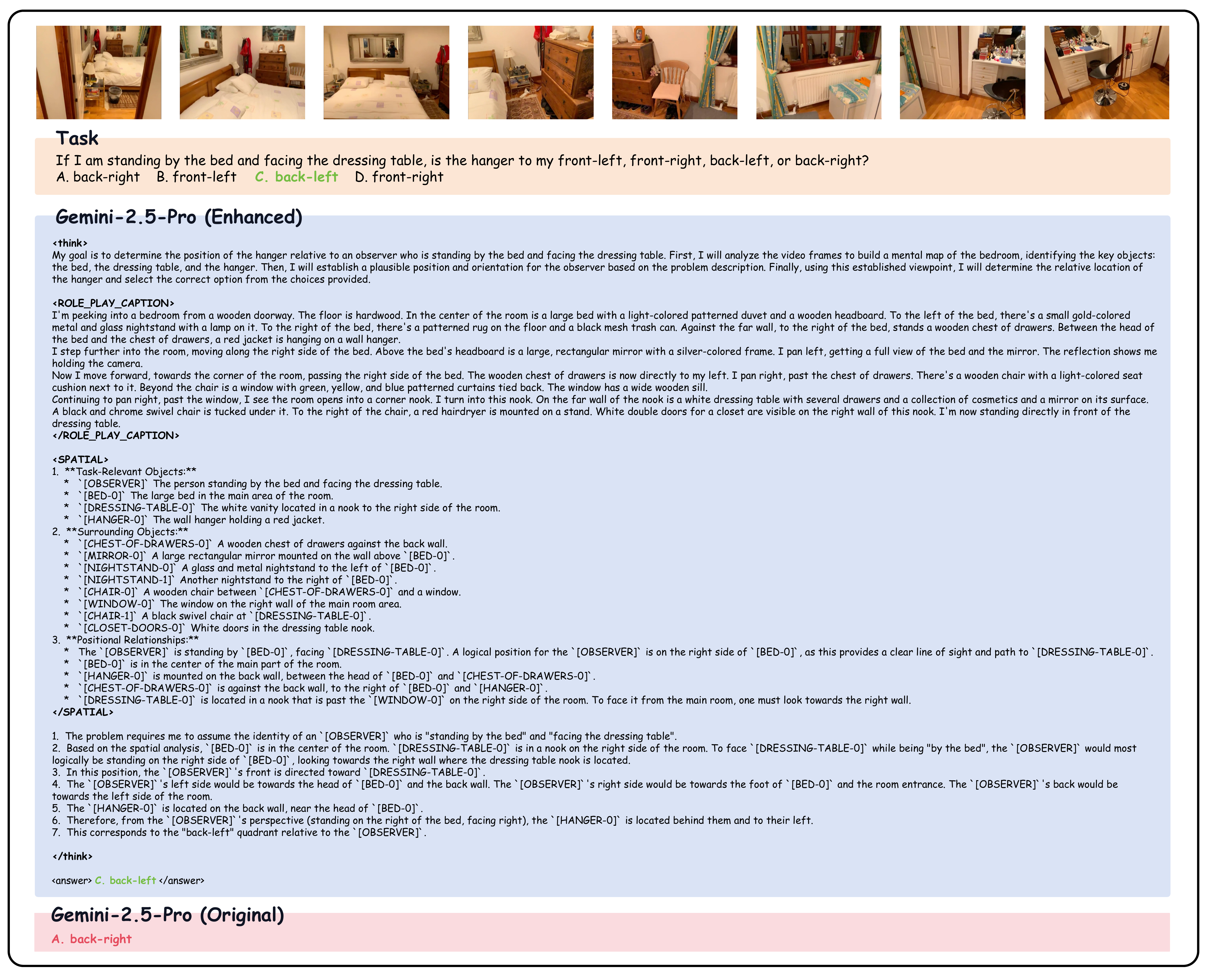}
    \vspace{-0.5cm}
    \caption{A case study where Gemini 2.5 Pro is guided by an EgoMind CoT prompt to solve a complex spatial relationship problem. This case illustrates that our proposed framework can be used as a zero-shot prompting strategy to unlock the spatial understanding and reasoning capabilities of powerful closed-source models.}
    \vspace{-0.3cm}
    \label{fig:gemini_case_en}
\end{figure*}

\section{Extended Ablation Studies}

\subsection{Generalization Across Model Scales} 

To evaluate the generalization ability of the EgoMind CoT across different MLLM scales, we fine-tune Qwen2.5-VL-3B using the proposed framework and assess its performance on VSI-Bench. The results are presented in Table~\ref{tab:VSI-Bench_qwen_ablation}.

As shown in Table~\ref{tab:VSI-Bench_qwen_ablation}, EgoMind yields consistent improvements on both Qwen2.5-VL-3B and Qwen2.5-VL-7B. With only the SFT stage, performance increases from 27.68 → 37.08 on the 3B backbone and from 30.02 → 42.33 on the 7B backbone, indicating that the EgoMind CoT effectively equips MLLMs with structured spatial reasoning abilities. When reinforcement learning is further introduced, the scores improve substantially to 45.44 and 50.16 for the 3B and 7B models, respectively, demonstrating the strong synergy between CoT-based supervision and RL-driven refinement.

It is worth noting that EgoMind requires only 5K automatically generated CoT samples for SFT and 20K QA-only samples for RL—significantly fewer than competing methods. These results highlight the high data efficiency of EgoMind and its ability to activate robust spatial cognition purely through carefully designed linguistic reasoning, without relying on additional multi--modal data or explicit 3D supervision.

\subsection{Intermediate Results Verification}
To rigorously verify the faithfulness and reliability of our generated reasoning chains, we employ Gemini 2.5 Pro as an independent judge to audit the intermediate reasoning traces on the VSI-Bench validation set. 

We evaluate the intermediate results across two key dimensions: (i) \textbf{Visual Fidelity}, which measures whether the generated RPC and PSA context accurately reflects the raw video frames; and (ii) \textbf{Logical Consistency}, which assesses whether the final answer logically stems from the reasoning chain. Specifically, we design detailed evaluation prompts that instruct the judge to assign a binary score (0 or 1) to each reasoning trace for both dimensions. These binary scores are then averaged to compute the final aggregate percentages. Our evaluation reveals that the EgoMind CoT achieves a high visual fidelity of \textbf{98.93\%} for RPC and \textbf{91.60\%} for PSA, alongside an impressive \textbf{96.69\%} logical consistency. This strong alignment between the intermediate reasoning steps and the final answer confirms that EgoMind's performance gains arise from reliable, grounded spatial perception, effectively mitigating the risk of spurious correlations or shortcut learning.

\begin{figure*}[!htbp]
    \centering
    \includegraphics[width=\linewidth]{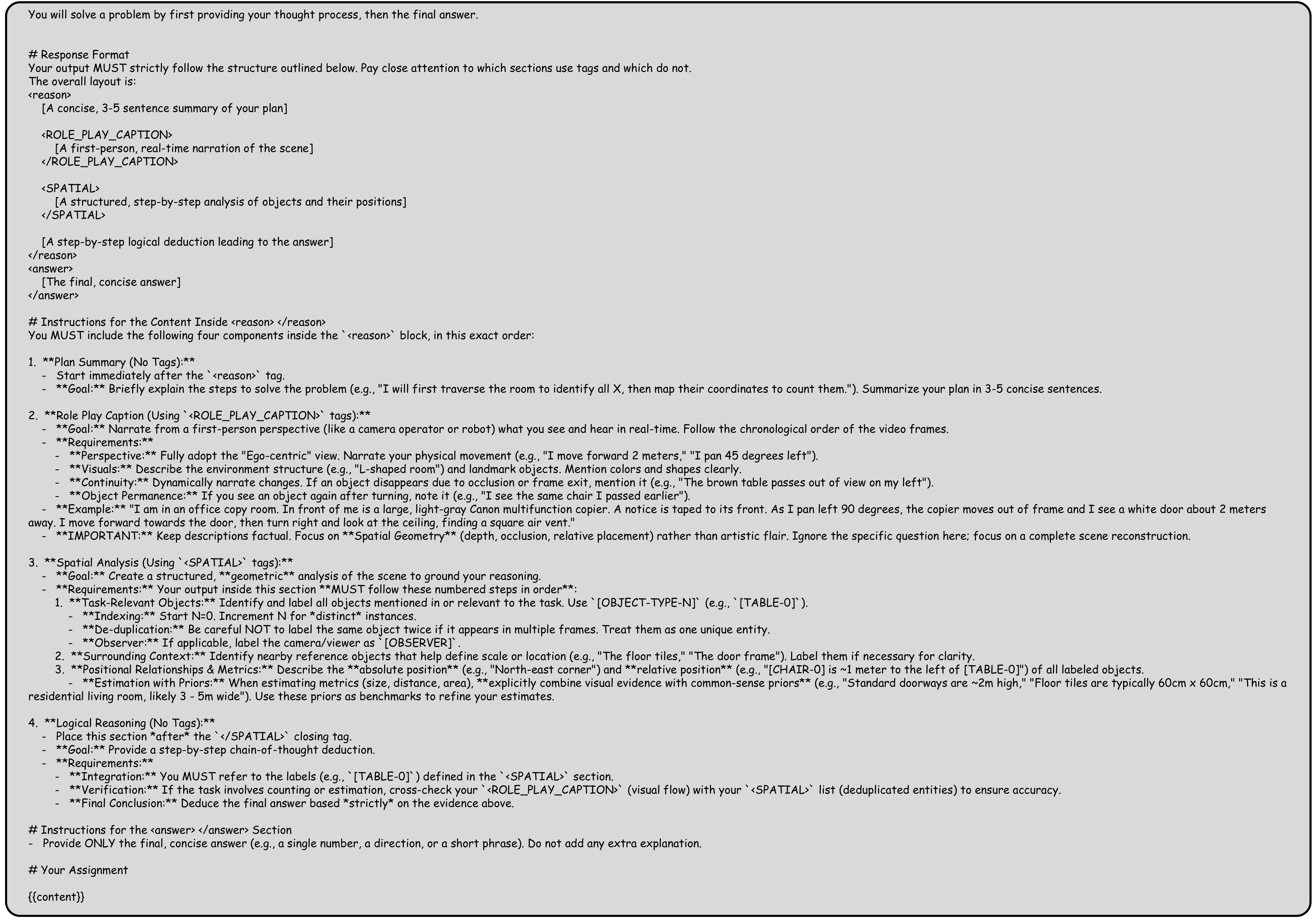}
    \caption{A zero-shot prompt to activate the spatial understanding and reasoning capabilities of powerful closed-source models.}
    \label{fig:zeroshot-prompt}
\end{figure*}

\section{Qualitative Analysis}
\label{sec:case_study}

To further investigate the qualitative improvements brought by EgoMind CoT, we evaluate the representative open-source MLLM, Qwen2.5-VL-7B, by comparing its responses with and without EgoMind-style reasoning. We categorize our qualitative analysis into relational understanding, metric consistency, and typical failure modes.

\noindent \textbf{Relational Reasoning Capabilities.} 
As visualized in Fig.~\ref{fig:qwen_case_en}, EgoMind CoT successfully activates robust spatial cognition in Qwen2.5-VL-7B for relative position and direction tasks. While the vanilla model struggles to maintain spatial awareness across multiple views, the enhanced model demonstrates the ability to construct a coherent, linguistically grounded spatial graph. It accurately identifies task-relevant objects across continuous frames and integrates these visual cues into a well-structured reasoning chain to deduce complex spatial relationships seamlessly.

\noindent \textbf{Insights on Metric Consistency.} 
Beyond qualitative relational reasoning, EgoMind excels at bridging semantic and metric information. Through the cross-frame alignment induced by the RPC and PSA modules, the framework enforces an implicit geometric consistency. As illustrated by the successful metric case in Fig.~\ref{fig:example}(a), this mechanism allows the model to maintain stable object identities and consistent scale cues across multiple viewpoints without the need for explicit 3D supervision. Consequently, EgoMind can more effectively leverage the implicit spatial priors inherent in MLLMs to support complex metric reasoning tasks, such as estimating room sizes or determining precise physical distances.

\noindent \textbf{Failure Cases and Extensibility.} 
Despite these strong spatial modeling capabilities, we identify two primary failure modes in highly complex scenarios. The first is \textit{anchor mismatch} shown in Fig.~\ref{fig:example}(b), which typically arises when environments contain multiple visually identical or similar objects, occasionally confusing the model's cross-frame object tracking. The second failure mode stems from \textit{abrupt perspective shifts} shown in Fig.~\ref{fig:example}(c), where severe or discontinuous camera movements lead to sparse visual anchors, breaking the coherent spatial narrative constructed by the RPC module. 

Nevertheless, the linguistic nature of EgoMind renders it highly extensible. While fine-grained metric precision can be challenging for pure 2D MLLMs, incorporating partial metric hints (e.g., basic object size cues) into the prompt can significantly mitigate these issues. In our exploratory experiments, providing such hints improved the \textit{Room Size} estimation accuracy on VSI-Bench from 40.35\% to 44.72\%, demonstrating the flexibility and adaptability of our CoT framework.

\section{Zero-Shot Performance}

Remarkably, even for Gemini 2.5 Pro—a closed-source model—the EgoMind zero-shot prompt (Fig.~\ref{fig:zeroshot-prompt}) elicits noticeably stronger spatial reasoning. As shown in Fig.~\ref{fig:gemini_case_en}, the guided reasoning structure enables Gemini 2.5 Pro to consistently capture cross-frame correspondences, recognize implicit spatial bridges, and assemble a more coherent global scene representation. Quantitatively, replacing the generic CoT prompt in Sec.~\ref{sec:training_strategy} with the EgoMind zero-shot prompt in Fig.~\ref{fig:zeroshot-prompt} further boosts Gemini 2.5 Pro on VSI-Bench from 50.62 to 59.73.

Conversely, applying the same zero-shot prompt to Qwen2.5-VL-7B yields only marginal gains from 30.02 to 32.89. This contrast reveals that while zero-shot CoT prompting alone can activate spatial reasoning in massive closed-source models, it is insufficient for smaller open-source models due to limited instruction-following capacities. Consequently, our two-stage training (SFT + RL) is indispensable for smaller models to fully internalize the reasoning paradigm, effectively driving performance to 50.16.

These findings demonstrate that EgoMind CoT is not bound to a specific model architecture. Instead, it serves as a generalizable and effective reasoning paradigm that substantially enhances spatial understanding in both open-source and closed-source MLLMs.


\end{document}